\documentclass[letterpaper, 10 pt, conference]{ieeeconf}  

\IEEEoverridecommandlockouts                              

\usepackage{graphicx}
\usepackage{multirow}
\usepackage{url}  
\usepackage{xcolor}
\usepackage{svg}
\usepackage{adjustbox}
\usepackage{tcolorbox}
\usepackage[colorlinks=true]{hyperref}

\let\labelindent\relax
\usepackage{enumitem}

\title{\LARGE \bf
Analyzing Human Perceptions of a MEDEVAC Robot in a Simulated Evacuation Scenario}

\author{Tyson Jordan, Pranav Pandey, Prashant Doshi, Ramviyas Parasuraman, Adam Goodie
\thanks{The authors are with the School of Computing (TJ, PP, PD, RP) and the Department of Psychology (AG) at the University of Georgia, Athens, GA 30602, USA. 
Author emails: {\small \{tysonjordan,pranav.pandey,ramviyas,pdoshi,goodie\}@uga.edu}}}

\begin{document}

\maketitle
\thispagestyle{empty}
\pagestyle{empty}

\begin{abstract}

The use of autonomous systems in medical evacuation (MEDEVAC) scenarios is promising, but existing implementations overlook key insights from human-robot interaction (HRI) research. Studies on human-machine teams demonstrate that human perceptions of a machine teammate are critical in governing the machine's performance. 
Here, we present a mixed factorial design to assess human perceptions of a MEDEVAC robot in a simulated evacuation scenario. Participants were assigned to the role of casualty (CAS) or bystander (BYS) and subjected to three within-subjects conditions based on the MEDEVAC robot's operating mode: autonomous-slow (AS), autonomous-fast (AF), and teleoperation (TO). During each trial, a MEDEVAC robot navigated an 11-meter path, acquiring a casualty and transporting them to an ambulance exchange point while avoiding an idle bystander. Following each trial, subjects completed a questionnaire measuring their emotional states, perceived safety, and social compatibility with the robot. Results indicate a consistent main effect of operating mode on reported emotional states and perceived safety. Pairwise analyses suggest that the employment of the AF operating mode negatively impacted perceptions along these dimensions. There were no persistent differences between casualty and bystander responses.


\end{abstract}
\section{Introduction}
Medical evacuation (MEDEVAC) entails acquiring wounded, injured, or ill persons and evacuating them to safety, often through hazardous environments. Conducting efficient MEDEVACs poses significant challenges in urban search and rescue \cite{murphy2017disaster} and military \cite{ATP4-02.2_2019} contexts. In some instances, casualties are provided en route care while transported to safety. Notwithstanding debate surrounding the importance of EMS response time \cite{NEWGARD2010235}, timeliness is often prioritized in these scenarios. Unfortunately, the urgency of MEDEVAC operations requires cursory decision-making and planning. Moreover, MEDEVACs often take place in hostile environments, which pose life-threatening risks for rescuers \cite{wilde2018user}. 

\begin{figure}[t]
    \centering
    \includegraphics[width=1\linewidth]{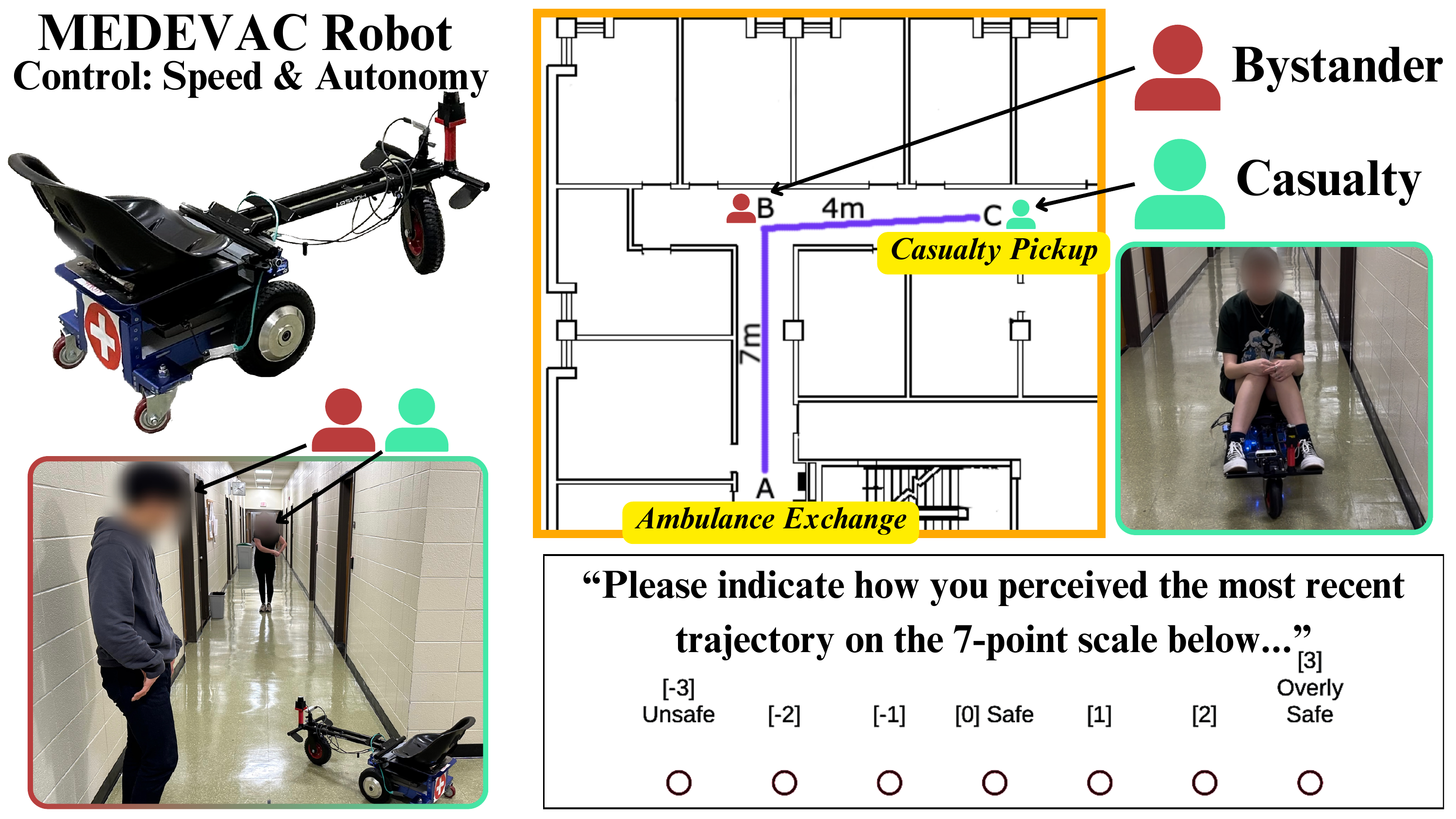}
    \caption{Our Simulated MEDEVAC Scenario. Participants assume the role of bystander or casualty (victim). They are subjected to several trials where a MEDEVAC robot is tasked with acquiring and evacuating the casualty from a casualty pickup point to a drop-off (ambulance exchange) point. Trials vary by the methods used to operate the robot. Following each trial, subjects complete a questionnaire designed to measure their perceived safety, emotional states, and social compatibility with regard to the robot. }
    \label{fig:overview}
    \vspace{-4mm}
\end{figure}

The introduction of autonomous systems in medical interventions has potential to improve the success of MEDEVAC operations. Amidst the numerous medical applications of autonomous systems \cite{Jeler}, the use of MEDEVAC robots in rescue has significant implications. Notably, deploying these systems for recovering casualties reduces the demand for emergency responders, demonstrating a more secure method of rescue. However, it is unclear how autonomous systems used for urban and military evacuations might be received by the rescuers employing them. Research in human-robot interaction (HRI) has highlighted the importance of human-robot compatibility in team-based efforts. The trust and acceptance of a machine by human teammates is a critical factor in determining the machine's performance \cite{groom}. 
Moreover, safety is a common consideration in developing robots intended for team settings. 
For MEDEVAC robots to perform optimally, it is crucial to maximize the compatibility between these systems and their users. This requires an understanding of the factors that influence perceptions of machine teammates.

To this end, our study borrows from several established multi-factor scales, such as the Godspeed (GS) and Robotic Social Attributes Scale (RoSAS), which are common measures in HRI designed to assess dimensions of perceived safety, trust, and general social compatibility \cite{Bartneck2009}, \cite{Carpinella}. Accordingly, we developed a MEDEVAC robot and explored its limitations 
in a controlled environment, measuring human perceptions of the robot. 
Our goal is to provide insights that will inform future design choices for casualty extraction \cite{Yoo2011}, significantly improving robot-aided rescue efforts.

In our experiment, human subjects were paired and assigned to one of two roles: 'casualty' or 'bystander.' Individual subjects maintained their role for the duration of the experiment. For each trial, a casualty subject awaited rescue as a bystander stood idle along a MEDEVAC robot's path. The robot navigated an 11-meter route to acquire and eventually transport the casualty to an ambulance exchange point. 
The robot had to avoid the bystander en route. Trials varied according to the robot's operating mode, which defined three primary within-subject conditions: autonomous-slow (AS), autonomous-fast (AF), and teleoperation (TO). These modes were characterized by the robot's speed and autonomy. After each trial, subjects completed a survey designed to measure their emotional states, perceptions of safety, and social compatibility with the robot. Our mixed factorial design allowed us to explore how subject role and the robot's operating mode influenced human perceptions along these dimensions. This unique study provides several key contributions to the field of HRI and to rescue robotics:
\begin{enumerate}[leftmargin=*,topsep=0in,itemsep=0in]
    \item We introduce the novel application of a MEDEVAC robot in a controlled experiment as a testbed for rescue scenarios, specifically measuring human subjects' emotional states, perceptions of safety, and social compatibility with respect to the robot. Moreover, this is the first study comparing these perceptions by differentiating the bystander and casualty roles, which are often intertwined in the literature \cite{humphrey2015human}.
    \item We employ a robust experimental framework to probe the factors influencing human perceptions of a MEDEVAC robot. Namely, we focus on the subjects' roles and the robot's operating modes. 
    \item Our results directly contribute to understanding how autonomous systems can be improved for medical intervention, offering practical implications that can guide the development of these systems in the future.
\end{enumerate}
\section{Related Work}
As autonomous systems expand into new sectors, there is a growing need to identify the factors that contribute to positive perceptions of machine teammates. Several efforts have been made to achieve a standardized measurement tool for HRI factors, but these scales typically differ by the dimensions they measure. One of the most widely used tools in this domain is the Godspeed (GS) scale, designed to measure general social perceptions of a machine teammate \cite{Bartneck2009}. 
GS focuses on five key concepts from HRI: animacy, anthropomorphism, likability, perceived intelligence, and perceived safety. 
In our study, we pay particular interest to GS's dimension of perceived safety. 
Before GS, methods emphasized perceived safety from the robot's perspective \cite{Murphy}. These efforts focused on the indirect measurement of affective states, \cite{Kuli2006EstimatingRI}, \cite{kulic}, whereas GS specifically measures human evaluations of a robot. 

Other scales stemming from GS, like the Robotic Social Attributes Scale (RoSAS) and Perceived Social Intelligence (PSI) scale, argue for the use of different dimensions in measuring social compatibility \cite{Carpinella}, \cite{barchard}. Unfortunately, neither the GS, RoSAS, or PSI have been held to the contemporary validation standards used in the organizational sciences \cite{schurer}. To address this shortcoming, we include the Human-Machine Teammate Inventory (HMTI) scale, which offers its own dimensions for assessing human perceptions of machine teammates \cite{grant_goodie_doshi_2024}. The HMTI has undergone rigorous 3-phase validation, positing task trustworthiness, cognitive understanding, explainability, liking, and empathy as factors of successful human-machine teams. Trust is a critical factor that impacts team performance \cite{costa}. HRI research has shown that higher-performing robots resulted in increased trust amongst human teammates \cite{rabby}. If human teammates perceive robots as untrustworthy, the robot's support is rejected, and its capacity to contribute to team-based outcomes is deprived \cite{groom}. In designing our questionnaire, we utilized numerous dimensions from the aforementioned measures.

Despite the abundance of social factors believed to influence human-machine teams, HRI has been largely overlooked in the development of MEDEVAC robots. Instead, research in rescue robotics has regularly placed emphasis on platform design and autonomous capabilities \cite{Murphy}. Notwithstanding the lack of attention paid to the compatibility of humans and these systems, there is existing work dedicated to improving the efficiency of autonomous systems in MEDEVAC scenarios \cite{delmerico}. For example, the European Land Robot Trial (ELROB) hosts a search-and-rescue MEDEVAC competition, where teams employ unmanned ground vehicles (UGVs) tasked with locating and evacuating dummy casualties \cite{schneider}, \cite{10018818}. The competition rewards speed and autonomy, whereas human intervention has resulted in penalties in previous years. The use of dummies for wounded persons further abstracts the simulation from real-world scenarios. For instance, the 1st-place winning team in the 2014 ELROB competition employed a "towing" strategy where their UGV dragged victims to safety \cite{Brüggemann2016}. While this approach was not penalized during the competition, it would prove impractical in actual rescue operations. Other robotics competitions have similarly promoted the advancement of autonomous rescue technologies \cite{casper,takahashi,Murphy2}. However, these contests have garnered criticism for their lack of HRI awareness \cite{drury}. Recent laboratory-based MEDEVAC simulations have emphasized safety, though these studies also use dummy casualties and thus do not measure human perceptions of machine teammates \cite{saputra}.

Looking towards more practical implementations, the Battlefield Extraction-Assist Robot (BEAR) was developed to extract wounded soldiers while navigating difficult terrain and withstanding enemy fire \cite{ruppert2010robots}. However, the BEAR 
did not possess autonomous capabilities and was controlled via teleoperation. 
Recent advances in rescue robotics include the Humanoid Rescue Robot for Calamity Response (HURCULES) \cite{choi_wonsuk}. HURCULES has semi-autonomous control, allowing navigation on uneven terrain during casualty extraction. Field experiments have demonstrated this newer model's effectiveness. Aside from physical experimentation, other studies focusing on robotic rescue intervention have tested novel MEDEVAC designs in physics engine simulators like Gazebo \cite{app11125414}.
Our study departs from these works by uniquely assessing the impact of a MEDEVAC robot's speed, autonomy, and subjects' roles on human perceptions. Understanding the influence of these factors is crucial, as human perceptions directly contribute to the success of human-machine teams.

\section{Methods}
This section outlines the experimental design, technical setup, measures, hypotheses, participants, procedure, and data analysis methods employed in our study.
\subsection{Experimental design}
We utilized a mixed factorial design to assess human perceptions of an autonomous MEDEVAC robot, focusing on two distinct roles and several key operating modes. 
\subsubsection{Subject's role (Between-subjects)}
Subjects were assigned the role of \textit{casualty} or \textit{bystander}. 
\begin{itemize}[leftmargin=*,itemsep=0in,topsep=0in]
    \item Bystander (BYS): Subjects assigned to the BYS role acted as idle observers of the MEDEVAC robot's trajectory. Bystanders did not interact with the robot directly, but served as a stationary obstacle planted between the ambulance exchange point and the casualty pickup point. This role is classically investigated in the HRI literature \cite{tsui}.
    \item Casualty (CAS): Subjects assigned to the CAS role were acquired by the MEDEVAC robot and transported to an ambulance exchange point. Casualties can be considered direct human teammates \cite{humphrey2015human}.
\end{itemize}

\subsubsection{Robot's operating mode (Within-subjects)}
The MEDEVAC robot possessed several operating modes characterized by its speed and autonomy for a given trial. 
\begin{itemize}[leftmargin=*,itemsep=0in,topsep=0in]
    \item Autonomous-Slow (AS): The MEDEVAC robot navigated autonomously at 0.3m/s.
    \item Autonomous-Fast (AF): The MEDEVAC robot navigated autonomously at 0.75m/s.
    \item Teleoperation (TO): A research assistant controlled the MEDEVAC robot unbeknownst to the participants, moving the robot at 0.5m/s. All trials were performed by the same teleoperator to maintain 
    consistency.
\end{itemize}
Subjects were exposed to each operating mode twice.

\subsection{Technical Setup}
We conducted real-world experiments using a Ubiquity Magni robot, engineered with a buggy attachment to imitate a MEDEVAC scenario. The robot, built with a robust aluminum chassis, can carry payloads up to 100 kg and features a differential drive motor that allows zero-degree (in-place) turns.
To tailor the robot for our study, we 
attached a commercially available Hover-1 Beast Buggy attachment to the Magni platform, enabling a person to sit on the robot during the experiment. The robot is equipped with a Hokuyo 2D LIDAR and an Intel Realsense RGB-D camera mounted at the front for perception and autonomy.

The robot's control system runs on ROS Noetic, managing the camera and LIDAR nodes. 
The FastSLAM algorithm \cite{bailey2006consistency} was used to perform localization and mapping by the robot. 
In autonomous navigation, we employed the $move\_base$ ROS package, which integrates a global and local planner to guide the robot to a designated goal while avoiding obstacles.
For global path planning, we utilized the $nav\_fn$ planner of $move\_base$, which leverages the $A*$ algorithm. This planner continuously recalculates the global path around any newly detected obstacles based on a specified frequency. For local path planning, we implemented the $teb\_local\_planner$ package. This planner optimizes the robot's trajectory in real time by balancing execution time, obstacle separation, and compliance with kinodynamic constraints using the Timed Elastic Band method. In the teleoperation mode, the operator used the real-time mapping output of the SLAM algorithm and live video feed to steer the robot.

\subsection{Measures}
Participants completed a questionnaire designed to assess their emotional states (Q1), perceived safety (Q2-Q3), and a variety of social judgments (Q4) with respect to the MEDEVAC robot and its performance. 
\begin{itemize}[leftmargin=*,topsep=0in,itemsep=0in]
    \item Q1: Question 1 consisted of five 5-point Likert-scale items measuring emotional states, including relaxation, calmness, quiescence, comfort, and predictability regarding their experience. Several of these factors were derived from the GS scale. 
    \item Q2: Question 2 was a single 7-point Likert-scale item measuring general perceived safety on a scale of 'Unsafe' (1) to 'Overly Safe' (7).
    \item Q3: Question 3 consisted of five 5-point Likert-scale items measuring levels of perceived safety at various points during the previous trial.
    \item Q4: Question 4 consisted of 16 5-point Likert-scale items designed to measure general social compatibility. Many of the items were derived from the GS, RoSAS, PSI, and HMTI scales.
\end{itemize}

The questionnaire\footnote{Available at \url{https://drive.google.com/file/d/1jnAh5lgrzYK1q6GqxNfYVGm9QopyF7U1/view?usp=sharing}} was trial-specific and thus completed by each subject six times. 

\subsection{Hypotheses}
We established numerous hypotheses dependant upon subjects' roles, the robot's operating modes, and the types of perceptions we intended to examine.

\begin{description}
    \item[H1:] Subjects will report significantly lower emotion state ratings when the AF mode is employed compared to the other operating modes.
    \item[H2:] CAS subjects will report significantly lower emotional state ratings compared to their BYS counterparts.
    \item[H3:] Subjects will report significantly lower perceived safety ratings when the AF mode is employed compared to the other operating modes.
    \item[H4:] CAS subjects will report significantly lower perceived safety ratings compared to their BYS counterparts.
    \item[H5:] Subjects will report significantly lower social compatibility ratings when the AF  mode is employed compared to the other operating modes.
    \item[H6:] CAS subjects will report significantly lower social compatibility ratings compared to their BYS counterparts.
\end{description}

In summary, we anticipate a more negative experience for casualty subjects and for all subjects when the AF mode is employed. Casualties being transported are directly in the robot's care, and higher speed might increase the risk of errors or collisions. 

\subsection{Participants}
Participants (N=61) of our IRB-approved study were recruited using the University of Georgia's Psychology Research Participant Pool. Undergraduate students taking an introductory psychology course are required to participate in a minimum amount of research hours to receive class credit. The study also included undergraduate Computer Science students, who participated without compensation. 47 of the 61 subjects identified as female (77.05\%), with participants averaging 20 years of age. 

\subsection{Procedure}
\subsubsection{Briefing}
Subjects were scheduled to attend the experiment in pairs. Upon arrival, subjects were provided with briefing forms outlining the experiment. Informed consent was obtained via signed consent forms. Participants were randomly assigned to either the bystander or casualty role. 

\subsubsection{Experiment}
Each pair of participants underwent six trials. Trials were distinguished by the robot's three operating modes. The three conditions were presented in a random order and then presented again in the same order. During each trial, the MEDEVAC robot traveled from an ambulance exchange point (Point A) to a casualty pickup point (Point C), passing the bystander at Point B. Upon reaching Point C, the robot turned around, stopped, and played an audio message instructing the casualty to sit. Once the casualty was seated, the robot returned to Point A, again passing the bystander. After reaching Point A, another audio message informed the casualty that it was safe to dismount. Following each trial, subjects completed a questionnaire. 

\subsubsection{Debriefing}
After completing all six trials of the experiment, subjects were debriefed, which included a thorough explanation of the study's goals and the different conditions that were tested. Subjects were also informed that a researcher controlled the robot remotely for the TO trials.

\subsubsection{Data Analysis}
Data were extracted and preprocessed. 
To align with the other items in the analysis, Q2 scores were scaled from a 7-point to a 5-point scale using the equation below.
$$
x_{5} = (x_{7}-1)(4/6) + 1 \eqno{(1)}
$$
For each subject, responses were averaged for each pair of trials with matching conditions (e.g., the first and second AS trials). This resulted in three responses per subject for each question. 

Following preprocessing, we conducted two-way mixed Analysis of Variance (ANOVA) statistical tests for questions 1—4. We chose the two-way mixed model for its ability to measure the influence of both within-subjects and between-subjects factors on a dependent variable, alongside any interaction effects \cite{fisher1970statistical}. In our analyses, subject role acted as the between-subjects factor and the robot's operating mode acted as the within-subjects factor. There is debate surrounding the use of parametric statistical tests on non-normal or ordinal data, but evidence suggests that these measures are more robust than nonparametric methods and that they can successfully be applied to Likert-style data or data that deviates from a normal distribution \cite{Sullivan2013-st}, \cite{norman2010likert}. However, we utilize parametric and nonparametric pairwise comparisons for a robust analysis. Importantly, we use the Wilcoxon test with the Kenward-Roger degrees-of-freedom method and Bonferroni adjustment for the pairwise comparisons of response values between roles and across operating modes. The Wilcoxon test is nonparametric and thus does not make assumptions about the distribution of its input data \cite{gibbons1993nonparametric}. We generated composite scores for each question by averaging the scores of their constituent Likert items \cite{rickards2012you}. 

\section{Results}
\begin{figure}[t]
   \centering
   \includesvg[width=1\linewidth]{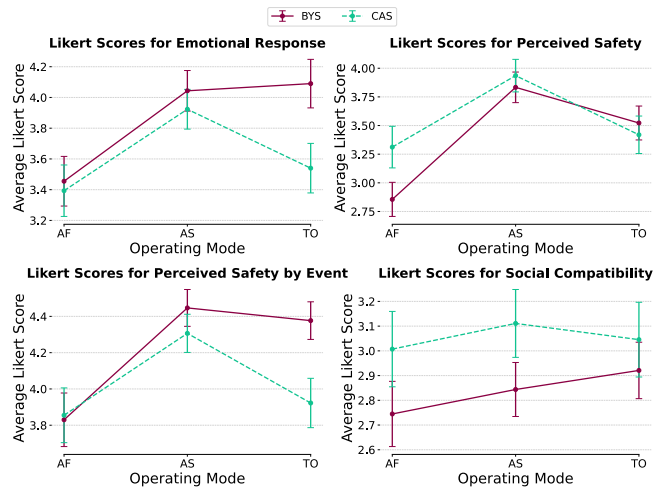}
   \caption{Average Likert-scale scores across subjects for Questions 1-4. }
   \label{fig:composite_score2}
   \vspace{-4mm}
\end{figure}

We report results from the statistical analyses of our Likert data. Results are organized by question. Average Likert response values for each question are displayed in Fig.~\ref{fig:composite_score2}. 

\subsection{Q1: Emotional State}
First, we analyzed Likert-scale ratings for Question 1, which was designed to gauge emotional states. A two-way mixed ANOVA was conducted to examine the effects of role and operating mode on composite score values for this question. We found a significant main effect of operating mode ($F(2, 116) = 20.222$, $p < .001$, $\eta^{2} = 0.259$). The interaction between role and operating mode was also significant ($F(2, 116) = 4.370$, $p = 0.015$, $\eta^{2} = 0.070$). However, the main effect of the role was not significant ($F(1, 58) = 1.674$, $p = 0.201$, $\eta^{2} = 0.028$). 

Pairwise comparisons of estimated marginal means revealed significantly lower scores within the AF operating mode compared to AS (estimate $= -0.559$, SE $= 0.090$, $t(116) = -6.198$, $p < .001$) and TO (estimate $= -0.391$, SE $= 0.090$, $t(116) = -4.332$, $p < 0.001$). The difference between AS and TO was not statistically significant (estimate $= 0.168$, SE $= 0.090$, $t(116) = 1.866$, $p = 0.194$). A Wilcoxon test confirmed a significant difference between scores for the AF and AS operating modes ($W = 1129$, $p < 0.001$, $p.\mathrm{adj} = 0.001$), confirming that the AF operating mode produced lower emotional state ratings than AS. The difference between AF and TO approached significance before adjustment ($W = 1396$, $p = 0.034$) but was not significant after applying Bonferroni correction ($p.\mathrm{adj} = 0.102$). A separate Wilcoxon test comparing scores within each role revealed that the AF operating mode was rated significantly lower than TO for bystander subjects ($W = 269$, $p = 0.008$, $p.\mathrm{adj} = 0.023$) but not for casualty subjects ($W = 420$, $p = 0.668$, $p.\mathrm{adj} = 1.000$). The Wilcoxon test revealed no significant difference between the AS and TO conditions ($W = 1982$, $p = 0.341$, $p.\mathrm{adj} = 1.000$).

We found no significant difference between bystander and casualty emotional states from our pairwise comparison tests. This is aligned with our mixed-ANOVA results. However, separating responses by operating mode revealed a significant difference between bystanders and casualties in the TO operating mode only (estimate $= 0.550, SE = 0.215$, $t(94.4) = 2.554$, $p = 0.0123$) ($W = 597$, $p = 0.030$), with casualties reporting significantly lower emotional states than bystanders within that condition.

\subsection{Q2: General Perceived Safety}
Next, we analyzed the effect of role and operating mode on general perceived safety. We again used a two-way mixed ANOVA to examine the effects of role and operating mode on response values. We found a significant main effect of operating mode ($F(2, 118) = 32.262$, $p < .001$, $\eta^{2} = 0.354$). The interaction between role and operating mode was also significant ($F(2, 118) = 4.025$, $p = 0.020$, $\eta^{2} = 0.064$), while the main effect of the role was not ($F(1, 59) = 0.674$, $p = 0.415$, $\eta^{2} = 0.011$). 

The pairwise comparisons of estimated marginal means revealed a significant difference in the perceived safety scores for the AF and AS operating modes, with AF scores being significantly lower (estimate $= -0.801$, SE $= 0.100$, $t(118) = -8.031$, $p < .001$). Similarly, reported perceived safety for AF was significantly lower than for TO (estimate $= -0.387$, SE $= 0.0997$, $t(118) = -3.883$, $p < 0.001$). In contrast, scores for the AS operating mode were significantly higher than TO (estimate $= 0.414$, SE $= 0.100$, $t(118) = 4.149$, $p < 0.001$). These findings were further supported by a Wilcoxon test, which indicated that AF scores were significantly lower than those for AS ($W = 976$, $p < 0.001$, $p.\mathrm{adj} < 0.001$). The comparison between AF and TO approached significance before adjustment ($W = 1418$, $p = 0.022$), but was not significant after applying the Bonferroni correction ($p.\mathrm{adj} = 0.068$). A follow-up comparison that separated responses by role revealed significantly lower perceived safety ratings for the AF operating mode compared to TO for bystanders ($W = 241$, $p = 0.002$, $p.\mathrm{adj} = 0.006$) but not casualties ($W = 466$, $p = 0.837$, $p.\mathrm{adj} = 1.000$). Lastly, the initial Wilcoxon test confirmed that AS scores were significantly higher than TO ($W = 2350$, $p = 0.012$, $p.\mathrm{adj} = 0.035$).

Pairwise comparisons of estimated marginal means and a Wilcoxon test revealed no significant difference between bystander and casualty perceived safety ratings when responses were averaged by operating mode, but did reveal significantly lower safety ratings from bystander subjects when the AF mode was isolated $t(106) = -2.094$, $p = .039$) ($W = 326$, $p = 0.043$).

\subsection{Q3: Perceived Safety by Event}
We then examined the effect of role and operating mode on perceived safety ratings by event. A two-way mixed ANOVA revealed a significant main effect of operating mode ($F(2, 118) = 23.939$, $p < .001$, $\eta^{2} = 0.289$). The interaction between role and operating mode was also significant ($F(2, 118) = 4.929$, $p = 0.009$, $\eta^{2} = 0.077$). However, the main effect of the role was not significant ($F(1, 59) = 1.511$, $p = 0.224$, $\eta^{2} = 0.025$). 

Pairwise comparisons of estimated marginal means revealed a significant difference in the perceived safety scores of the AF and AS operating modes, with AF generating significantly lower scores than AS (estimate $= -0.534$, SE $= 0.076$, $t(118) = -6.893$, $p < .001$). Similarly, AF scores were significantly lower than those for TO (estimate $= -0.307$, SE $= 0.076$, $t(118) = -3.965$, $p < .001$). In contrast, the scores for the AS mode were significantly higher than those for TO (estimate $= 0.227$, SE $= 0.076$, $t(118) = 2.929$, $p = 0.012$). Similar to Question 2, a follow-up Wilcoxon test confirmed a significant difference between the AF and AS conditions ($W = 1159$, $p < 0.001$, $p.\mathrm{adj} < 0.001$), indicating that AF scores were significantly lower than those for AS. The comparison between AF and TO approached significance before adjustment ($W = 1454$, $p = 0.037$), but was not significant after applying Bonferroni correction ($p.\mathrm{adj} = 0.111$). An auxiliary comparison of scores where responses were divided by role revealed significantly lower perceived safety ratings for the AF condition compared to TO for bystanders ($W = 262$, $p = 0.005$, $p.\mathrm{adj} = 0.016$) but not casualties ($W = 456$, $p = 0.740$, $p.\mathrm{adj} = 1.000$). The Wilcoxon test did not suggest a significant difference between AS and TO scores ($W = 2184$, $p = 0.097$, $p.\mathrm{adj} = 0.290$). 

Pairwise comparisons of estimated marginal means and a Wilcoxon test revealed no significant difference between bystander and casualty responses when averaged by operating mode, but did reveal significantly lower safety ratings from casualty subjects for TO only ($t(99.6) = 2.544$, $p = 0.013$) ($W = 624$, $p = .021$).

\subsection{Q4: Social Compatibility}
Finally, we inspected responses for any effect of role or operating mode on reported social compatibility. A mixed ANOVA indicated a significant main effect of operating mode ($F(2, 118) = 4.621$, $p = 0.012$, $\eta^{2} = 0.073$). However, the main effect of the role was not significant ($F(1, 59) = 1.401$, $p = 0.241$, $\eta^{2} = 0.023$). The interaction between role and condition was also not significant ($F(2, 118) = 2.084$, $p = 0.129$, $\eta^{2} = 0.034$).

Pairwise comparisons of estimated marginal means revealed significant differences in the social compatibility ratings between the AF and AS conditions, with AF generating significantly lower social compatibility ratings than AS (estimate $= -0.101$, SE $= 0.040$, $t(118) = -2.557$, $p = 0.035$). Similarly, scores for AF were significantly lower than those for TO (estimate $= -0.107$, SE $= 0.040$, $t(118) = -2.702$, $p = .024$). There was no significant difference between the AS and TO scores (estimate $= -0.006$, SE $= 0.040$, $t(118) = -0.146$, $p = 1.000$). Despite these findings, a Wilcoxon test with Bonferroni correction found no significant difference between any of the operating modes. Additionally, neither pairwise comparison method revealed a significant difference in social compatibility reports between roles.

\section{Discussion of Results}
The results of the study provide insights into human perceptions of a MEDEVAC robot in a controlled simulation. The mixed factorial design allowed us to explore several hypotheses regarding the effects of subject role and the robot's operating mode on subjects' perceptions of the robot.

\subsection{Emotional State (H1 \& H2)}
Hypotheses 1 and 2 pertained to subjects' emotional states, which were measured in Q1 of our questionnaire. Specifically, H1 posited that subjects would report significantly lower emotional state ratings when the AF operating mode was employed compared to the other modes. This is supported by our findings, suggesting that the robot’s higher speed may induce discomfort or stress for participants. H2 predicted that casualties would report lower emotional states compared to their bystander counterparts. While Q1 ratings were consistent between bystanders and casualties for the autonomous operating modes, they varied significantly for TO, as casualty ratings were significantly lower. The act of being evacuated by the MEDEVAC robot when it was operated remotely by a human negatively impacted casualties' emotional states. Notably, none of the composite scores fell below 3, the average score on our 5-point Likert scale; reported attitudes were positive-leaning across all roles and operating modes.

\subsection{Perceived Safety (H3 \& H4)}
Hypotheses 3 and 4 regarded perceived safety. We measured general perceived safety in Q2 and perceived safety by event in Q3. H3 predicted lower perceived safety ratings for the AF operating mode compared to the other two modes. This aligns with the results from Questions 2 and 3, with the exception that AF and TO did not engender significantly different ratings from the casualty subjects. This characterizes a discrepancy between bystander and casualty responses: Bystanders reported significantly lower general perceived safety (Q2) when the AF operating mode was employed, while casualties reported significantly lower perceived safety by event (Q3) when the TO operating mode was employed. Notably, bystanders had visual access to the MEDEVAC robot for the entire duration of each trial, whereas the casualties did not see the robot until it reached point B. It is possible that hasty or erratic behaviors occurring during the robot's route from point A to point B negatively impacted bystanders' perceptions of safety. This would not influence responses to Q3, which only measures perceived safety starting at the robot's trajectory from point B. Importantly, Q2 and Q3 utilized different Likert scales—For Q2, subjects reported their general perceived safety on a scale from Unsafe (1) to Overly Safe (7), while Q3 asked subjects to rate whether several events of the previous trial were safe or not, on a scale of Strongly Disagree (1) to Strongly Agree (5).

\subsection{Social Compatibility (H5 \& H6)}
Hypotheses 5 and 6 made predictions about reported social compatibility, which was measured by Q4. Specifically, H5 suggested that subjects would report lower social compatibility ratings when subjected to the AF condition. While our mixed ANOVA and pairwise comparisons of estimated marginal means supported this, our Wilcoxon test did not corroborate this finding. Because of this discrepancy, the results of our parametric analyses for Q4 reports should be considered with caution. 
Question 4 yielded sub-average composite scores for bystander subjects across each condition. As items in Q4 were selected from numerous scales, some might not have easily applied to our MEDEVAC scenario. H6 predicted that casualty subjects would report lower social compatibility scores compared to bystanders, but there was no significant difference between the reports of these groups for the social compatibility items.

\subsection{Limitations}
This study is subject to several limitations that should be acknowledged. First, there is no accepted universal scale for measuring social compatibility, which posed a challenge in developing Question 4. Question 4 contained 16 items, some of which might not have applied to our particular study. The inclusion of extraneous items might have contributed to the statistical discrepancies between our parametric and nonparametric methods.

\section{Conclusion} 
In this study, we conducted a rigorous human-machine team experiment, emulating a controlled evacuation scenario with a real-world robot. We assessed how a MEDEVAC robot's autonomy, speed, and the role of subjects (bystander or casualty) influenced subjects' perceptions. Our analysis revealed that subjects' emotional states and perceived safety were consistently more negative when the robot operated autonomously at a faster speed. In contrast, the employment of a slower autonomous mode resulted in more positive reports along these dimensions. There were no systematic differences between casualty and bystander responses. Instead, the variance between these groups' responses was nuanced and intermittent across the questionnaire. These findings emphasize the importance of carefully considering operating speed in developing autonomous systems, especially for MEDEVAC and rescue applications. Future designs should prioritize maintaining positive human perceptions without sacrificing safety or operational efficiency.

\section*{Acknowledgements}
Research was sponsored by the DEVCOM Analysis Center and was accomplished under Cooperative Agreement Number W911NF-22-2-0001. The views and conclusions contained in this document are those of the authors and should not be interpreted as representing the official policies, either expressed or implied, of the DEVCOM Analysis Center or the U.S. Government. The U.S. Government is authorized to reproduce and distribute reprints for Government purposes notwithstanding any copyright notation herein.

\bibliographystyle{IEEEtran}
\bibliography{root}

\end{document}